\documentclass[11pt, reqno]{amsart}

\usepackage{graphicx}
\usepackage{subfigure}
\usepackage{float}
\usepackage{url}            
\usepackage{booktabs}       
\usepackage{amsfonts}       
\usepackage{nicefrac}       
\usepackage{microtype}      
\usepackage{xcolor}         
\usepackage[utf8]{inputenc} 
\usepackage[T1]{fontenc}    
\usepackage[colorlinks=true, allcolors=blue]{hyperref}     
\usepackage{mathrsfs}
\usepackage{amsmath}
\usepackage{amssymb}
\usepackage{mathtools}
\usepackage{amsthm}
\RequirePackage[linesnumbered,ruled,vlined]{algorithm2e}
\usepackage[capitalize,noabbrev]{cleveref}
\usepackage[margin=2.5cm]{geometry}
\usepackage[textsize=tiny]{todonotes}
\usepackage{textcase}

\newcommand{\grad}{\mathrm{grad}}

\newcommand{\maxpool}{\mathrm{maxpooling}}
\newcommand{\mean}{\mathrm{mean}}

\newcommand{\interior}{\mathrm{int}}
\newcommand{\x}{{\bf{x}}}
\newcommand{\bu}{{\bf{u}}}
\newcommand{\ctrl}{\mathrm{ctrl}}
\newcommand{\divergence}{\mathrm{div}}

\newcommand{\odesolve}{{\mathrm{ODESolve}}}
\newcommand{\bignode}{{BigNode }}
\newcommand{\bigoc}{{BigOC }}

\theoremstyle{plain}
\newtheorem{theorem}{Theorem}[section]

\theoremstyle{definition}
\newtheorem{definition}[theorem]{Definition}

\theoremstyle{remark}

\makeatletter
\def\@evenhead{\rlap{\footnotesize\thepage}\hfil}
\makeatother

\title[Inverse Boundary Value and Optimal Control Problems on Graphs]{Inverse Boundary Value and Optimal Control Problems on Graphs: A Neural and Numerical Synthesis}

\author{Mehdi Garrousian\textsuperscript{*}
        \and
        Amirhossein Nouranizadeh\textsuperscript{*}
        }

\begin{document}

\maketitle

{
    \renewcommand{\thefootnote}{}
    \footnotetext{* Equal contribution. Emails: \MakeLowercase{\texttt{mgarrous@alumni.uwo.ca}}, \MakeLowercase{\texttt{nouranizadeh@aut.ac.ir}}.}
}

\thispagestyle{empty}

\begin{abstract}
A general setup for deterministic system identification problems on graphs with Dirichlet and Neumann boundary conditions is introduced. When control nodes are available along the boundary, we apply a discretize-then-optimize method to estimate an optimal control. A key piece in the present architecture is our boundary injected message passing neural network. This will produce more accurate predictions that are considerably more stable in proximity of the boundary. Also, a regularization technique based on graphical distance is introduced that helps with stabilizing the predictions at nodes far from the boundary. 
\end{abstract}

\section{Introduction}

The intersection of neural networks, system identification, and control has seen significant advancements in recent years. The universal approximation power of neural networks \cite{cybenko1989approximation, hornik1991approximation} makes them well-suited for data-driven modeling of dynamical systems. This, in turn, highlights the suitability of neural differential equation models for capturing complex patterns in diverse systems.

In this work, we are interested in the question of system identification on a graph with boundary and related optimal control problems. In this context, graphs serve as discrete topological spaces that support evolving scalar and vector fields over time. A scalar/vector field is a map that assigns a value to each node/edge of the graph. Our fundamental assumption is that the values of adjacent nodes interact, and that the system's evolution over time is, in a deterministic way, expressed by an unknown differential equation.

Throughout this manuscript, we work with graphs that have interior as distinct from boundary nodes. The interior nodes are subject to the internal dynamics of the system, whereas boundary nodes are exposed to an external source of dynamics. Moreover, in the context of a control problem, there are also control nodes that we can change for a cost. The training data consists of the scalar field values at a sequence of timestamps, and our goal is to learn the underlying dynamics well enough to perform inference and optimal control. Specifically, we are interested in the following problems:

\begin{enumerate}
 \item A \emph{Graphical System identification Problem} is about learning the interior dynamics of an unknown system on a graph which is exposed to known external dynamics through its boundary nodes/edges. This type is also known as an inverse problem. 
 \item A \emph{Graphical Optimal Control Problem} is about controlling the (identified) interior dynamics of a graph through its control nodes/edges given known dynamics on its boundary and subject to an objective and control cost. 
 \end{enumerate}
 We will call our solutions to these questions \emph{Boundary Injected Graph Neural ODE} and \emph{Boundary Injected Graphical Optimal Control}, for short \bignode and BigOC, respectively. Our techniques draw from message passing graph neural networks, neural ODE solvers and direct numerical optimal control in the discrete nonlinear case. Apart from the synthesis of methods and algorithms, our innovation lies in the treatment of exterior dynamics. In the literature, it is well understood that boundary values in the spatial dimension tend to be hard to learn \cite{chen2020comparison, lyu2020enforcing}. The standard approach is the penalty method which attempts to leverage the loss function to account for both the internal dynamics and the dynamics around the boundary. In contrast, we take a more structured approach where we allow the topology of the space to play a more decisive role and let the learnable differential operator be informed directly by the boundary values. 

Our two neural network components, namely graph neural networks and neural ODEs, are generalizations of convolutional nets and recurrent networks, respectively \cite{chen2018neural, gcn, graphsage}. In the latter context, teacher forcing is a standard technique that helps with stabilizing the learning process. In a manner that mimics the teacher forcing technique, we employ a \emph{boundary injection} technique in the message passing network that proves natural and helpful for our setup. 

These techniques are applied to diffusion phenomenons in both linear and nonlinear cases. This is due to the importance and ubiquity of these processes in various domains such as physics, viral propagations, marketing understanding, technology adoption, various financial processes, etc \cite{bass1969new, mahajan1990new, centola2010spread, carcione2020simulation}. 
 
The present work is adjacent to many of these papers in this domain which makes precise attributions and comparisons difficult. As a result, we are only going to attempt to mention a small subset of the relevant literature. In terms of core architecture, we are building on the foundations of \cite{zhou2020graph} on GNNs and \cite{chen2018neural} on neural ODEs, among others. In terms of control theory, although not used here, we found the treatment of \cite{jin2020pontryagin} to be interesting for its comprehensiveness. Other comparable approaches can be found in \cite{drgovna2022differentiable, asikis2022neural}. Throughout, we use physical laws to generate synthetic data which gives us a point of comparison with the literature on physics informed neural networks such as \cite{raissi2017physics}. Our experiments are only about diffusion processes, in both linear and nonlinear cases, which makes our work relevant to the line of research around \cite{chamberlain2021grand}. We give a careful treatment of the boundary values on graphs. This was also a central topic in \cite{lyu2020enforcing}. 

Here is a summary. In Section \ref{conceptsection}, we introduce the concept of a graph with boundary which is natural for questions where an external source influences the interior of a system. Then, in Section \ref{exp1}, we show with empirical results that our model \bignode produces remarkably stable and accurate inferences on random systems. Building on these foundations, in Section \ref{sec:goc}, we develop an offline discrete optimal control technique which is subsequently applied to an experiment in Section \ref{exp2}.

\section{Concepts}\label{conceptsection}
Our system identification model \bignode is a regression model consisting of a boundary injected GNN wrapped in a neural ODE with a graphical distance regularizer. Mathematically, this is an integral $ \int_{0}^{t} D_{\Theta}(\x_{\interior G}(\tau), \x_{\partial G}(\tau)) d\tau$. Let us unpack the pieces.
\subsection{Graphs and operators}
\begin{definition}\label{graphwithbound}
 A graph with boundary $G=(V_{int}, V_{\partial}, E)$ is a connected simple graph consisting of distinct interior $V_{int}$ and boundary nodes $V_{\partial}$ where boundary nodes can only connect to interior nodes.
 
 In the context of a control problem, we have two distinct types of boundary nodes: externally determined boundary nodes as above, as well as control nodes $V_{ctrl}$ which can take arbitrary values at a cost. To be explicit, we may write $G=(V_{int}, V_{\partial}, V_{ctrl}, E)$. 
 \end{definition}

\begin{definition}
 A scalar field on a graph is a map $\x:V_G\to {\mathbb{R}}$. Also, a vector field is a map ${\mathbf{X}}:E_G\to{\mathbb{R}}$. If $H\subset G$ is a subgraph, then $\x_{H}$ and ${\mathbf{X}}_H$ will be the restrictions to the nodes and edges of $H$, respectively.  
\end{definition}
For convenience, we may identify scalar/vector fields with the vectorized forms, e.g., $\x=(x_i)_{i\in V}$ where $x_i = \x(i)$. 
\begin{definition}
Consider a scalar field $\x$ with value $x_i$ at any node $i$. Along any edge $ij\in E$, the gradient $\grad\, \x$ makes a vector field as defined by:
\begin{equation}\label{eqn:grad}
 (\grad\, \x)({i, j}) = x_j - x_i.
\end{equation} 
Almost dually, given a vector field ${\mathbf{X}}$, its divergence $\divergence {\mathbf{X}}$ is a scalar field defined by:
\begin{equation}
 (\divergence {\mathbf{X}})(i) = \sum_{j\in \mathscr{N}_i} {\mathbf{X}}(i, j)
\end{equation}
where $\mathscr{N}_i$ is the collection of neighbors of node $i$. 

Using these operators, one can defined the graph Laplacian operator $\Delta$ (also denoted: $L$, $\Delta_0$) equivalently as:
\begin{equation}
\Delta = \grad^*\grad = - \divergence\, \grad
\end{equation}
or more explicitly as $\Delta = D-A$ where $D$ is the diagonal matrix of degrees and, $A$ is the adjacency matrix. 
\end{definition}
Diffusion on a graph is typically described by
\begin{equation}\label{linear_heat}
 \dot{\x} = - \kappa \Delta \x
\end{equation}
where $\kappa$ is a coefficient of diffusitivity. 

If it is believed that the rate of diffusion changes with the value of the scalar field, then a more general model may be required, such as:
\begin{equation}\label{nonlinear_heat}
 \dot{\x} = \divergence (k(\x) (\grad\,\x)).
\end{equation}
The Wiedemann–Franz law describes typical choices for $k$, in the context of thermal and electrical conductivity, such as linear and exponential functions of the form:
\begin{equation}\label{conductivities}
 k_1(x) = k_0 + a(x-x_0), \quad k_2(x) = k_0 (x/x_0)^n
\end{equation}
where $k_0$ is the diffusion rate at base value $x_0$, and $n$ is a suitable power. 

\subsection{Dirichlet, Neumann and control problems}

We are interested in learning temporal scalar fields on graphs that are governed by unknown dynamics. First, let us take a step back to clarify the context. A typical deterministic Dirichlet initial value problem on a graph with boundary is of the form
 \begin{equation}
  \begin{cases}
     \dot{\x}_{\interior G}(t)   =  {\mathrm{D}}(\x(t), t) \\
    \x_{\interior G}(0)  {\textrm{ given}}\\
    \x_{\partial G}(t)  {\textrm{ given for all }t}
  \end{cases}  
 \end{equation}
where $\mathrm{D}$ is a differential operator (e.g., any type of graph Laplacian) that expresses the time derivative of $\x$ in terms of its values on the interior, boundary and time. Moreover, the initial value of $\x$ over the interior and the boundary values at all times are given, and time $t$ belongs to an interval starting from zero. Alternatively, when the third constraint is replaced with ${(\grad\,\x)_{\partial G}}$ for all $t$, then we call it a Neumann type problem. Note that our treatment of the interior is distinct from the boundary as the boundary is subject to an external source of dynamics. 

In contrast, we are interested in graphical system identification problems which are inverse to the above type in the sense that the differential operator is unknown. Our approach is to approximate $\mathrm{D}$ as a neural network $\mathrm{D}_{\Theta}$ by allowing it to learn from a dataset generated by the latent system at timestamps $T=\{0, t_1, \dots, t_N \}$.  

\begin{definition}\label{inverse_dirichlet}
 An inverse Dirichlet problem on a graph $G$ is about learning an unknown differential operator ${\mathrm{D}}_{\Theta}$ and integrating a scalar field $\x$ on the interior of $G$ subject to a system of the form
 \begin{equation}
  \begin{cases}
     \dot{\x}_{\interior G}(t)   =  {\mathrm{D}}_{\Theta}(\x(t), t) \\
    \x_{\interior G}(0)  {\textrm{ given}}\\
    \x_{\partial G}(t)  {\textrm{ given for all }}t 
  \end{cases} 
 \end{equation} 
 by using training data consiststing of a set of the form
\[
 \{ (t_i, \x(t_i)_{\interior G},  \x(t_i)_{\partial G}): t_i\in T\}
\]
where $T$ is a finite set of observed timestamps, and $\x(t)_{\partial G}$ is the external scalar field on the boundary nodes at time $t$.
\end{definition}

Note that the statement of the problem requires the boundary values of all $t$ within the range while training data only provides these values at specific timestamps. In order to fill the intermediate values, we interpolate linearly between consecutive boundary values. 

\begin{definition}
 Similarly, an inverse Neumann problem on a graph $G$ is about learning ${\mathrm{D}}_{\Theta}$ and integrating $\x$ subject to:
 \begin{equation}
  \begin{cases}
    \dot{\x}_{\interior G}(t)  =  {\mathrm{D}}_{\Theta}(\x(t), t) \\
    \x_{\interior G}(0)  {\textrm{ given}}\\
    {(\grad\,\x)_{\partial G}}(t) {\textrm{ given for all }}t
  \end{cases}  
 \end{equation}
  In this case, the training data consist of 
  \[
 \{ (t_i, \x(t_i)_{\interior G},  (\grad\,\x)(t_i)_{{\partial G}}): t_i\in T\}, 
\]
where $(\grad\,\x)(t)_{{\partial G}}$ is the gradient flow on the boundary edges at time $t$. 
\end{definition}

It is natural to have Robin boundary problems that have a mixture of the two types of boundary values. Next, let us take a look at a typical graphical control problem. In the following definition, the operator $D$ is identified whether analytically or as a learned neural network. 

\begin{definition}
 A (Dirichlet) graphical optimal control problem seeks to find an optimal scalar field $\bu(t)$ on the control nodes $V_{\ctrl}$ subject to a system:
\begin{equation}\label{ctrl_problem}
  \begin{cases}
     \dot{\x}_{\interior G}(t)   =  {\mathrm{D}}(\x(t), t) \\
    \x_{\interior G}(0)  {\textrm{ given}}\\
    \x_{\partial G}(t)  {\textrm{ given for all }}t \\
    \x_{\ctrl}(t) = \bu(t)
  \end{cases}     
\end{equation}
such that the internal trajectory $\x_{\interior G}(t)$ is close enough to a desired internal state $\hat{\x}_{\interior G}(t)$ and with a small enough terminal error $E_{\textrm{term}}$ and optimal relative to an objective function 
\begin{equation}\label{continuous_objective}
 J  = \int_0^{t_N}   \alpha||\x_{\interior G}(t)-\hat{\x}_{\interior G}(t)||^2+ \beta ||\bu(t)||^2 dt  + \gamma E_{\textrm{term}}(\x_{\interior G}(t_N), \hat{\x}_{\interior G}(t_N)). 
\end{equation}
where $\alpha$, $\beta$ and $\gamma$ depend on the context, and $||\cdot ||$ stand for the $L_2$-norm or an appropriate variant. 
\end{definition}
Similarly, one can define the Neumann version of the above OC problem. Also, the objective function may be modified to accommodate more general requirements. In our setting, we are most interested in the case where $D=D_{\Theta}$ is an identified neural network. 

\section{Message passing with boundary injections}\label{neural_layers}

The differential operator $D_{\Theta}$ as a graph neural network ${\rm{D}}_{\Theta} = {\rm{D}}^{(\ell)}\cdots {\rm{D}}^{(1)}$ is a composition of $\ell$ message passing layers where each ${\rm{D}}^{(k)}$ has a message function $\phi^{(k)}$ and an update function $\gamma^{(k)}$. 
\begin{center}
\begin{figure*}[t]
  \centering
\includegraphics[scale=0.09]{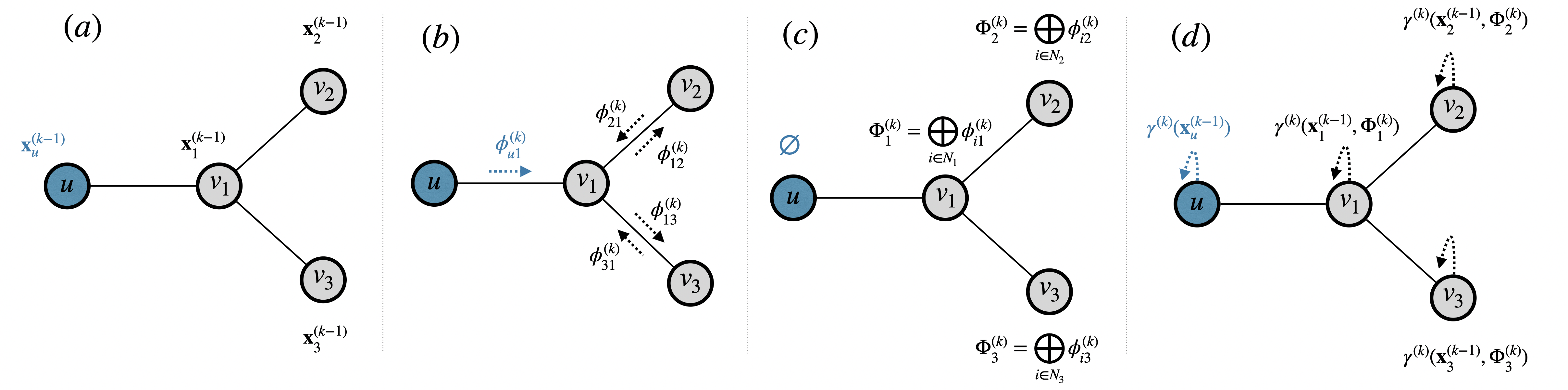} 
  \label{fig:mp}
  \caption{A visualization of boundary injected message passing as in Formula (\ref{msg_passing}) where node $u$ is boundary/control and gray nodes are interior. Plot (b) shows the flow of messages. Plot (c) shows the aggregation step. Plot (d) shows the update process.}
\end{figure*}
\end{center}
For any node $i\in G$ and any $k=1, \dots, \ell$, we have
\begin{equation}\label{msg_passing}
  x_i^{(k)} = (D^{(k)}\x^{(k)})_i =  
  \gamma^{(k)}\left( x_i^{(k-1)}, \bigoplus_{j\in \mathscr{N}_i} \phi^{(k)}(x_i^{(k-1)}, x_j^{(k-1)}, e_{ji}) \right).
\end{equation}
When $i \in \interior G$ is an interior node, $\mathscr{N}_i$ consists of its neighbors whether from the interior or outside. The edge case: when $i \in \partial G$ (or $\in \ctrl G$ in case of a control problem), we let $\mathscr{N}_i$ be empty. This is due to Definition \ref{graphwithbound}, and the assumption that we are in an open system where the effects of the interior toward the outside are negligible, as is typical in thermodynamics. Effectively, in the edge case, this will put a zero tensor in place of the direct sum. Also, the direct sum is an aggregation operator such as $\mean, \maxpool$, etc. In more general cases, we may appropriately assign edge weights. 

Dimensionwise, we have $x_i^{(k)}\in {\mathbb{R}}^{d^{(k)}}$ where
\begin{equation}\label{embed_dims}
 {d^{(k)}} = \begin{cases} \mathtt{input\_dim} & k = 0 \\ \mathtt{embed\_dim} & 0< k < \ell \\ \mathtt{input\_dim} & k = \ell\end{cases}.
\end{equation}
The message function $\phi^{(k)}$ maps from $2 d^{(k)}$ dimensions (along with optional edge features) to $\mathtt{message\_dim}$. In our experiments, we used $\mathtt{embed\_dim}= \mathtt{message\_dim} =32$. The update function $\gamma^{(k)}$ maps from $d^{(k-1)}+m^{(k)}$ dimensions to $d^{(k)}$ dimensions, and may carry an activation function such as softplus, etc.

In practice, we have been using the following concrete form. For $i \in \interior G$, we have
\begin{equation}\label{msg_passing_int}
 x_i^{(k)} = {\mathrm{L}}_{self}^{(k-1)} x_i^{(k-1)} + {\mathrm{L}}_{ngbr}^{(k)} \frac{1}{|\mathscr{N}_i|}\sum_{\substack{j\in \mathscr{N}_i}} {\mathrm{L}}_{msg}^{}(x_i^{(k-1)}, x_j^{(k-1)}) 
\end{equation}
where mean aggregation is used, ${\mathrm{L}}_{self}, {\mathrm{L}}_{ngbr}, {\mathrm{L}}_{msg}$ are single dense layers of correct dimension with appropriate activation functions, and the last pair is concatenation. 

If $j \in \partial G$ (or $\in \ctrl G$), then the second term gets dropped:
\begin{equation}\label{msg_passing_bound}
 x_j^{(k)} = {\mathrm{L}}_{self}^{(k-1)} x_j^{(k-1)}. 
\end{equation}
During training, $\x^{(0)}_{\partial G}$ is available at all times either as boundary values or interpolated values between them. The transformations in (\ref{msg_passing_bound}) take place independently from the interior values. Such boundary values will be consumed in (\ref{msg_passing_int}) whenever $j\in \partial G$. See Figure \ref{msg_passing}. This modified message passing architecture is an adaptation of the teacher forcing technique in language models which we call \emph{boundary injection}. 
\subsection{Regularization}\label{section:regularization}
\begin{figure}[htbp]
    \centering
\includegraphics[scale=0.33]{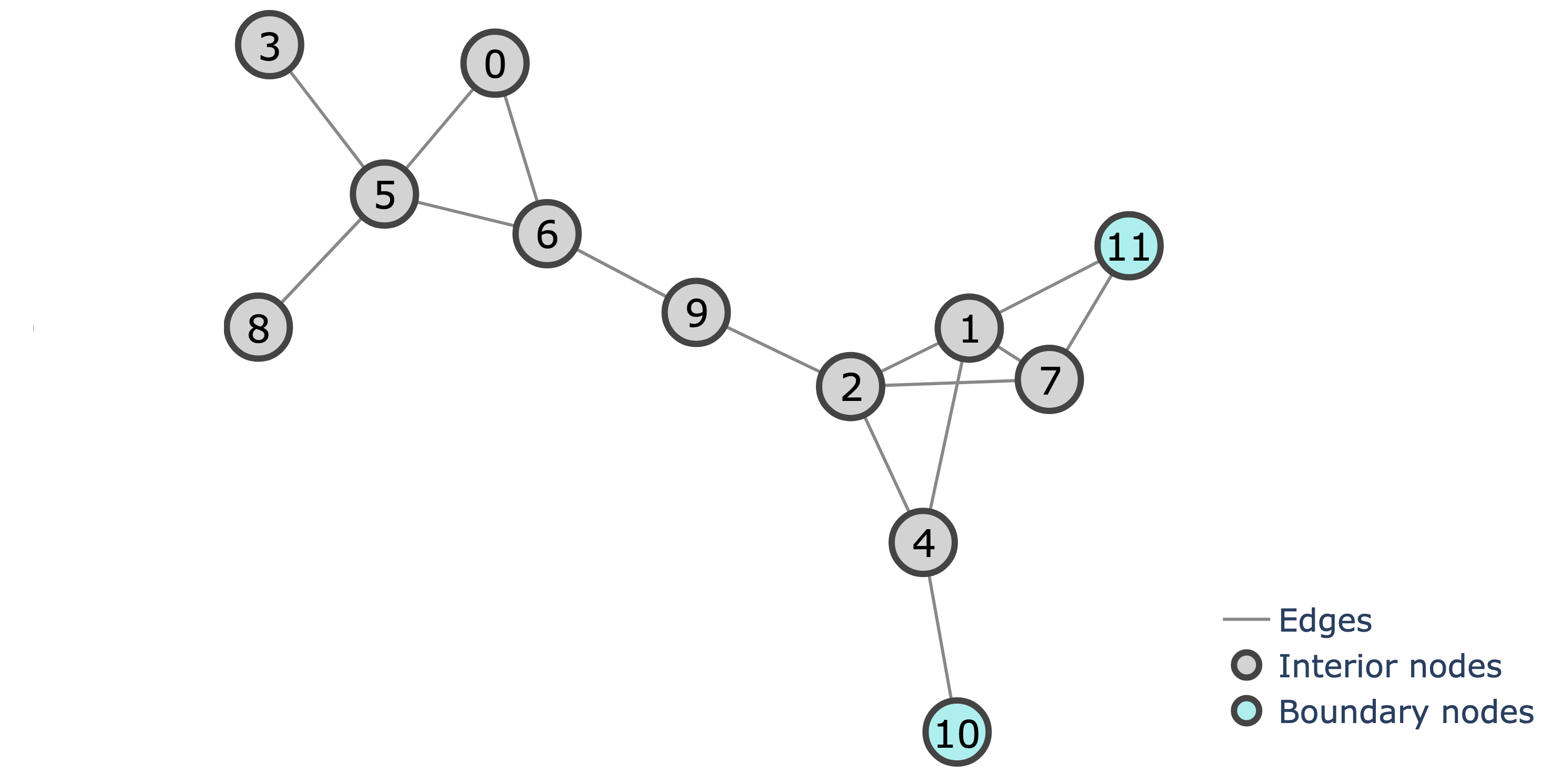}
\caption{Graph $G_1$ used in SysID experiments}
\label{exp1_plot}
\end{figure}

Nodes that are near the sources of heat train much easier and more accurately than nodes that are far from them. This phenomenon calls for a regularization technique that would attend more to the farther nodes. A simple recipe that we found to be effective is based on the distance of the interior nodes to the closest boundary nodes. 
\begin{equation}
 d_i = \min \{ d(i, j): j\in \partial G\}
\end{equation}
For graph $G_1$ (Figure \ref{exp1_plot}), the vector of distances is $(5, 1, 2, 6, 1, 5, 4, 1, 6, 3)$. Then, we modify the MSE loss by applying weights to bump up the penalty of wrong predictions at farther nodes.
\begin{equation}
 Loss(\x_{\interior G}, \hat{\x}_{\interior G}) = \sum_{i \in \interior G} w_i (x_{i} - \hat{x}_i)^2
\end{equation}
The general rule is that if $d_i > d_j$, then $w_i \geq w_j$. Beyond that, we treat the weights as hyperparametrs. In practice, we noticed that even weights that assign $w_i = 2$ to the farthest nodes and else $w_i=1$ were effective in combatting the instability of learning at those far nodes. 

\section{Experiment 1: graphical system identification}\label{exp1}

In this section, we are going to apply \bignode to two system identification problems, one linear and one nonlinear, both supported on the same random graph $G_1$ as shown below with two boundary nodes and ten interior nodes. Throughout these experiments, we work with a fixed random graph and random boundary values unless stated otherwise. 
We generate synthetic data by integrating the diffusion equations as discussed in Section \ref{graphwithbound} in both linear and nonlinear cases. For the linear example, we use Euler's method to integrate Equation \ref{linear_heat} with $\Delta t = 10^{-3}$, $\kappa = 10$, $N=1000$ steps, and polynomial and sine boundaries. The \bignode model consists of a GNN as in Section \ref{neural_layers} with $\ell = 2$, and with fully connected linear layers as message and update functions of appropriate dimensions. For training, we use an Adam optimizer, the MSE loss, and train for 2000 epochs. As a baseline, we use a plain GNN as differential operator wrapped in a neural ODE (named Gnode) but without any boundary injection. See Subsection \ref{add_plots} for a complete set of plots. 

For our nonlinear experiment, again, we use the Euler method to integrate the nonlinear diffusion formula \ref{nonlinear_heat} with thermal conductivity of type $k_1$ as in \ref{conductivities} with $a=10, k_0=x_0=0$, and $\Delta t = 0.001$, $N=1000$. The boundary nodes 10 and 11 have the following trajectories. 

\begin{center}
\includegraphics[scale=0.063]{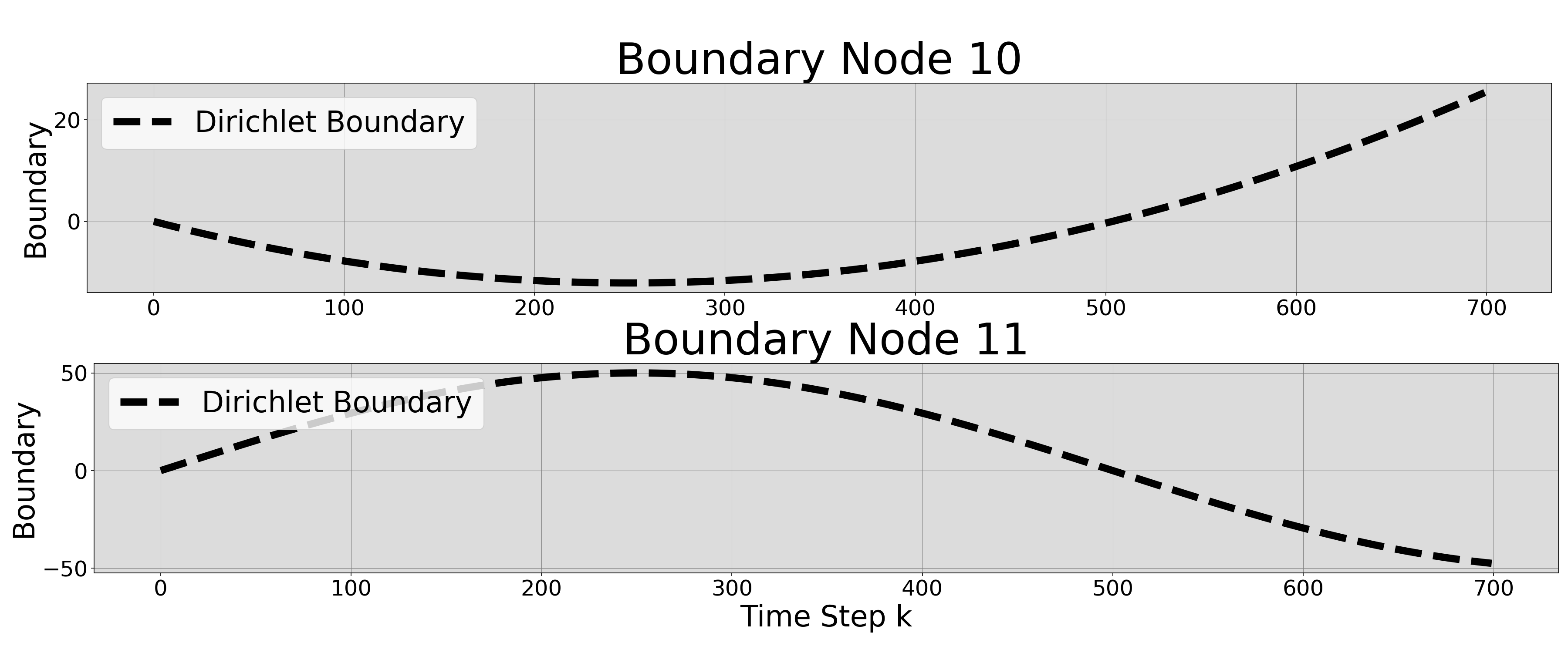}
\end{center}
Our observation is that nodes closer to the boundary receive stronger signals which allows the model to learn their dynamics better. For instance, at node 2, both standard \bignode and regularized \bignode do reasonably well. 

\begin{center}
\includegraphics[scale=0.26]{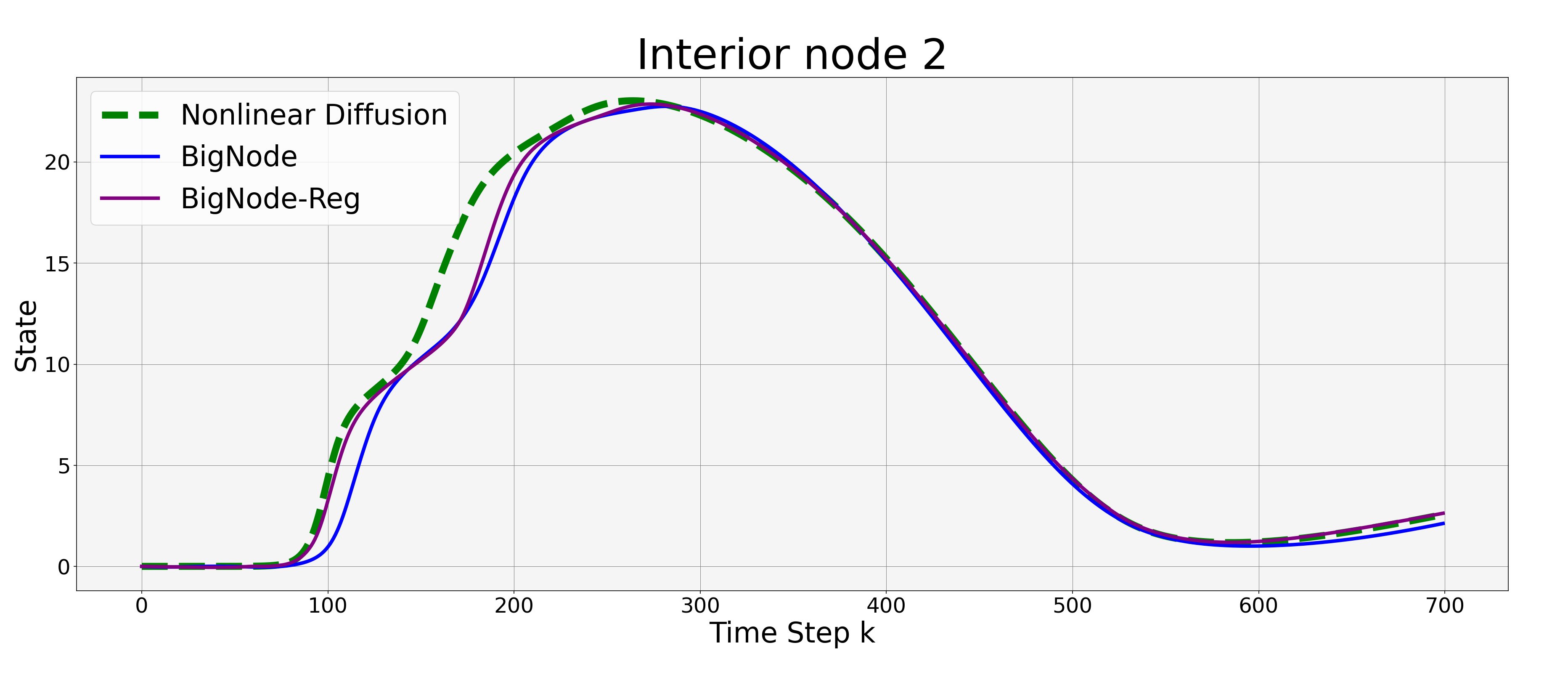}  
\end{center}
In contrast, learning at the nodes farthest from the boundary proves to be considerably more challenging. Luckily, our regularization technique from Section \ref{section:regularization} can improve learning efficiency. For instace, see node 0.

\begin{center}
\includegraphics[scale=0.26]{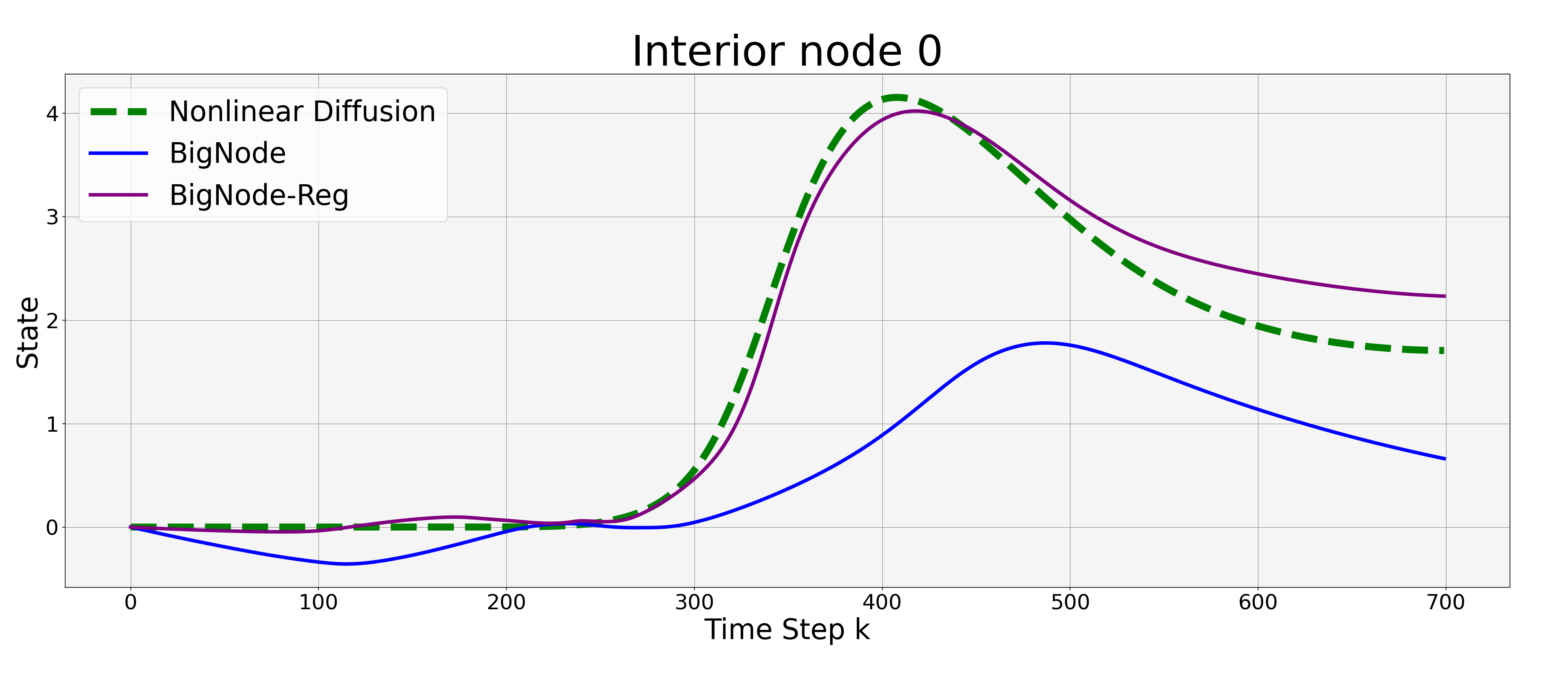} 
\end{center}

\begin{table}[h]
    \centering
    \begin{tabular}{lcccc}
        \toprule
        & \multicolumn{2}{c}{Train} & \multicolumn{2}{c}{Test} \\
        \cmidrule(lr){2-3} \cmidrule(lr){4-5}
       RMSE & Mean & Std & Mean & Std \\
        \midrule
        Gnode & 12.629 & 6.136 & 20.636 & 21.733 \\
        \bignode & 0.108 & 0.013 & 0.699 & 0.173 \\
        \bignode Reg & 0.104 & 0.007 & 0.639 & 0.172 \\
        \bottomrule
    \end{tabular}
    \caption{RMSE values for a linear system}
    \label{tab:linear_stacked}
\end{table}

\begin{table}[h]
    \centering
    \begin{tabular}{lcccc}
        \toprule
        & \multicolumn{2}{c}{Train} & \multicolumn{2}{c}{Test} \\
        \cmidrule(lr){2-3} \cmidrule(lr){4-5}
       RMSE & Mean & Std & Mean & Std \\
        \midrule
        Gnode & 9.54 & 4.512 & 6.679 & 5.421 \\
        \bignode & 1.326 & 0.579 & 0.414 & 0.138 \\
        \bignode Reg & 0.167 & 0.144 & 0.302 & 0.13 \\
        \bottomrule
    \end{tabular}
    \caption{RMSE values for a nonlinear system}
    \label{tab:nonlinear_stacked}
\end{table}

\section{Graphical optimal control}\label{sec:goc}

In this section, alongside boundary nodes, our graphs are also equipped with control nodes $V_{\ctrl}$. With a differential operator $D_{\Theta}$ possessing ample generalization capabilities, our objective is to derive optimal control values corresponding to a predefined objective. Our overarching strategy involves discretization followed by optimization as facilitated by discrete optimal control tools. See \cite{betts2010practical} for further details. To achieve this, we express the optimality conditions of the problem (\ref{ctrl_problem}) as a Lagrangian and employ an iterative Newton-like method for its resolution.

Originally, the objective (\ref{continuous_objective}) is an integral. In order to reduce it to a finite problem, we are going to discretize it as 
\begin{equation}
 J = \sum_{k=1}^{N-1} L(\x_{\interior G}(t_k), \x_{\ctrl G}(t_k))  + \gamma E(\x_{\interior G}(t_N)).
\end{equation}
To simplify notation, we'll use the shorthands $\x_k = \x_{\interior G}(t_k)$, ${\mathbf{b}}_k={\x_{\partial G}}(t_k)$, and $\bu_k = \x_{\ctrl G}(t_k)$. Also, we are going to omit $\x_{\partial G}$ Throughout. Now, the optimization loss can be written as
\begin{equation}\label{opt_loss}
 J = \sum_{k=0}^{N-1} L(\x_k, \bu_k)  + \gamma E(\x_{N})
\end{equation}
where $L(\x_k, \bu_k) = \alpha ||\x_k - \hat{\x}_k||^2+\beta||\bu_k||^2$, and $\hat{\x}_k$ is the desired value. The system dynamics is determined by
\begin{equation}\label{odesolve_update}
\x_{k+1} = \x_k + \odesolve(\x_k, {\mathbf{b}}_k, \bu_k, D_{\Theta}, t_k, t_{k+1})
\end{equation}
for $k=0\dots, N-1$, as we integrate incrementally. Here, $\odesolve(\x_k, {\mathbf{b}}_k, \bu_k, D_{\Theta}, t_k, t_{k+1})$ is a partial \bignode (a neural ODE evaluation) which is essentially the integral curve $ \int_{t_k}^{t_{k+1}} D_{\Theta}(\x(t), \bu(t)) dt$, 
starting at $(\x_k, \bu_k)$. Let us denote the right hand side of (\ref{odesolve_update}) as $g(\x_k, \bu_k)$, suppressing the time inputs for conveneince. Next, we will follow the recipe of \cite{diehl2011numerical} to derive the Karush-Kuhn-Tucker (KKT) equations. The data of the system dynamics with prescribed initial value can be packed into a matrix 
\begin{equation}
 G(\x , \bu) = \left[\begin{array}{c} g(\x_0, \bu_0) - \x_1 \\ g(\x_1, \bu_1) - \x_2 \\ \vdots \\ g(\x_{N-1}, \bu_{N-1}) - \x_N  \\ \x_0-\bar{\x}_0\end{array}\right].
\end{equation}
We have a Lagrangian
\begin{equation}
\begin{aligned}\label{eqn:lagrangian}
  \mathcal{L}(\x, \bu, \lambda) & = J + \lambda^T G \\
  & = J + \sum_{k=0}^{N-1} \lambda_{k+1}(g(\x_k, \bu_k)-\x_{k+1})\\ 
  & +\lambda_r(\x_0-\bar{\x}_0). 
\end{aligned}
\end{equation}
Assuming no constraint on the terminal value, the above system boils down to: 
\begin{equation}
 \begin{aligned}
 \nabla_{\x, \bu} \mathcal{L}(\x, \bu, \lambda)=0 \\ 
 G(\x, \bu) = 0 \\ 
 \x_0 = \bar{\x}_0 
 \end{aligned}
\end{equation}
This system can be solved simultaneously, as stated above, or sequentially, which comes down to three steps, namely a forward sweep to compute the state $\x$, a backward sweep to find the costates $\lambda$, and a residual equation to find the optimal values $\bu$. Let's expand these steps. 
\begin{itemize}
 \item The forward sweep accounts for the dynamics. Given the initial value $\bar{\x}_0$,
 \begin{equation}\label{forward_eqns}
   \x_{k+1} = g(\x_k, \bu_k), 
 \end{equation}
 for $k=0, \dots, N-1$.  
 \item The backward sweep accounts for the gradient of the Lagrangian wrt $\x_k$.
 \begin{equation}\label{lambdas} 
  \begin{aligned} \lambda_N & = \nabla_{\x_N} E(\x_N) \\   
  \lambda_k & = \nabla_{\x_k} L(\x_k, \bu_k) + \frac{\partial g}{\partial \x_k}(\x_k, \bu_k)^T \lambda_{k+1}, 
  \end{aligned}
 \end{equation}
 for $k=N-1, \dots, 1$. 
 \item The residuals account for the gradient of the Lagrangian wrt $\bu_k$. 
 \begin{equation}
  \nabla_{\bu_k} L(\x_k, \bu_k)+ \frac{\partial g}{\partial \bu_k}(\x_k, \bu_k)^T \lambda_{k+1} = 0,  
 \end{equation}
 for $k=0, \dots, N-1$. 
\end{itemize}
Most gradients are straightforward to compute with automatic differentiation except for the partial derivatives $\partial g/\partial \x_k$ and $\partial g/\partial \bu_k$ which pose more challenges. Set ${\mathbf{U}} = [\bu_0, \dots, \bu_{N-1}]^T$. Using equations (\ref{forward_eqns}) and (\ref{lambdas}), we can eliminate all $\x_k$ and $\lambda_k$ from the residual equations and write them in the form $R({\mathbf{U}})=0$, then we employ a Newton scheme
\begin{equation}\label{newtons}
 ({\mathbf{U}})_{new} = {\mathbf{U}} - (J_{R}({\mathbf{U}}))^{-1}R({\mathbf{U}})
\end{equation}
to iteratively solve for ${\mathbf{U}}$. Alternatively, we may use a approximate method such as BFGS as sketched in \ref{bfgs}. Next, we are going to outline a method to tackle the partial derivatives. An alternative will be given based on Runge-Kutta 4 in \ref{rk4}. 

\subsection{Variations with sensitivities}

We have
\begin{equation}
 g(t, \x_k, \bu_k) = \x_k + \int_{t_k}^{t_k+t} D_{\Theta} (\x(\tau), \bu(\tau)) d\tau
\end{equation}
for $0\leq t \leq t_{k+1}-t_k$, where we are interested in $G(t)=\partial g/\partial \x_k$ and $H(t) = \partial g/\partial \bu_k$. The trick is to first compute $\dot{G}$ and $\dot{H}$ and then integrate to recover the original functions. We have
\begin{equation}
\begin{aligned}
  \dot{G}(t) & = \frac{d}{dt}\frac{\partial g}{\partial \x_k} = \frac{\partial }{\partial \x_k}\frac{\partial g}{\partial t}= \frac{\partial }{\partial \x_k}D_{\Theta}(g, \bu) \\
  & =\frac{\partial D_{\Theta}}{\partial \x} \frac{\partial g}{\partial \x_k} =\frac{\partial D_{\Theta}}{\partial \x_k} G(t)
\end{aligned}
\end{equation}
where the order of differentiation is switched and the chain rule is applied. Note that the last partial derivative is efficiently computed using automatic differentiation. This produces an auxiliary differential equation
\begin{equation}
 d \ln(G(t)) = \frac{\partial D_{\Theta}}{\partial \x_k}dt, \quad G(0)=I. 
\end{equation}
Similarly, for $H$, we get an auxiliary equation
\begin{equation}
\begin{aligned}
  \dot{H} & = \frac{\partial}{\partial \bu_k} D_{\Theta}(\x(t), \bu(t)) = \frac{\partial D_{\Theta}}{\partial \x_k} H(t) + \frac{\partial D_{\Theta}}{\partial \bu_k} \frac{\partial \bu(t)}{\partial \bu_k}\\
  & = \frac{\partial D_{\Theta}}{\partial \x_k}H(t)+\frac{\partial D_{\Theta}}{\partial \bu_k}, 
\end{aligned}
\end{equation}
with $H(0) = 0$. In total, we get $2N$ differential equations, one per each index $k=0, \dots, N-1$ of each type above. These equations can be solved alongside a system identification problem using a neural ODE solver. Having solved these, we plug the results back into the residual equations and proceed. 

\subsection{Control algorithm}

Note that Sensitivities and RK4 (\ref{rk4}) are alternative methods to tackle the hard derivatives required in KKT. The former is more accurate but requires a more costly compute. In contrast, RK4 is a much faster approximate method. When assuming that the differential operator is linear, all instances of $G$ for $k=0, \dots, N-1$ turn out to be the same value. This allows us to save one order of magnitude. In general though, the incremental RK4 proved to be a reasonable approximate alternative. 

Iterating on the residual equations using conventional Newton's method can be hindered by the computationally expensive inversion of the Jacobian. In contrast, BFGS has demonstrated itself as a swift and efficient compromise.

When given a Neumann boundary condition, we use the following formula to convert it back to Dirichlet type. If $j \in V_{\ctrl G}$, $i\in \interior G$, and $e=ji \in E$, then
\begin{equation}
    \x_j  = \x_i - (\grad\, \x)(j, i) 
\end{equation}
When dealing with multiple edges connecting across, we take the sum. Lastly, note that if boundary values are presented, an extension of the above algorithm with modified forward sweep will work. 
\begin{algorithm}\label{main_algo}
\caption{Graphical Optimal Control \bigoc}
\KwData{$\x_{\interior G}, \x_{\partial G}$, along with multiple control scenarios $\x_{\ctrl G}^{(1)}, \dots, \x_{\ctrl G}^{(s)}$ at times $t_0, \dots, t_N$}
Extend $\x_{\partial G}$ by linear interpolation\;
\If{has Neumann boundary values}{convert to Dirichlet boundary values}
Set up graph distance regularization (Section \ref{section:regularization})\;
Set up message passing layers (Section \ref{neural_layers})\;
\For{each $\x_{\ctrl G}^{(1)}, \dots, \x_{\ctrl G}^{(s)}$}{train \bignode;
}
{\textbf{Ensure}} The resulting SysId model generalizes well to unseen inputs\;
\While{norm of gradient of Residual system not small enough}{
    Set up KKT system (Formula \ref{eqn:lagrangian})\;
    Run forward pass to get $\x_k$\;
    Run backward pass to get $\lambda_k$\;
    Compute $\nabla_{\x}\mathcal{L}$ using Sensitivities or RK4\;
    Run a Newton or BFGS iteration to solve for $\bu_k$\ for all $k$;
}
\KwResult{Trained neural ODE network
\;
Optimal control $\x_{\ctrl G} = \bu$ subject to the costs\;
}
\end{algorithm}

\section{Experiment 2: graphical optimal control}\label{exp2}
\begin{figure}[htbp]
    \centering
\includegraphics[scale=0.33]{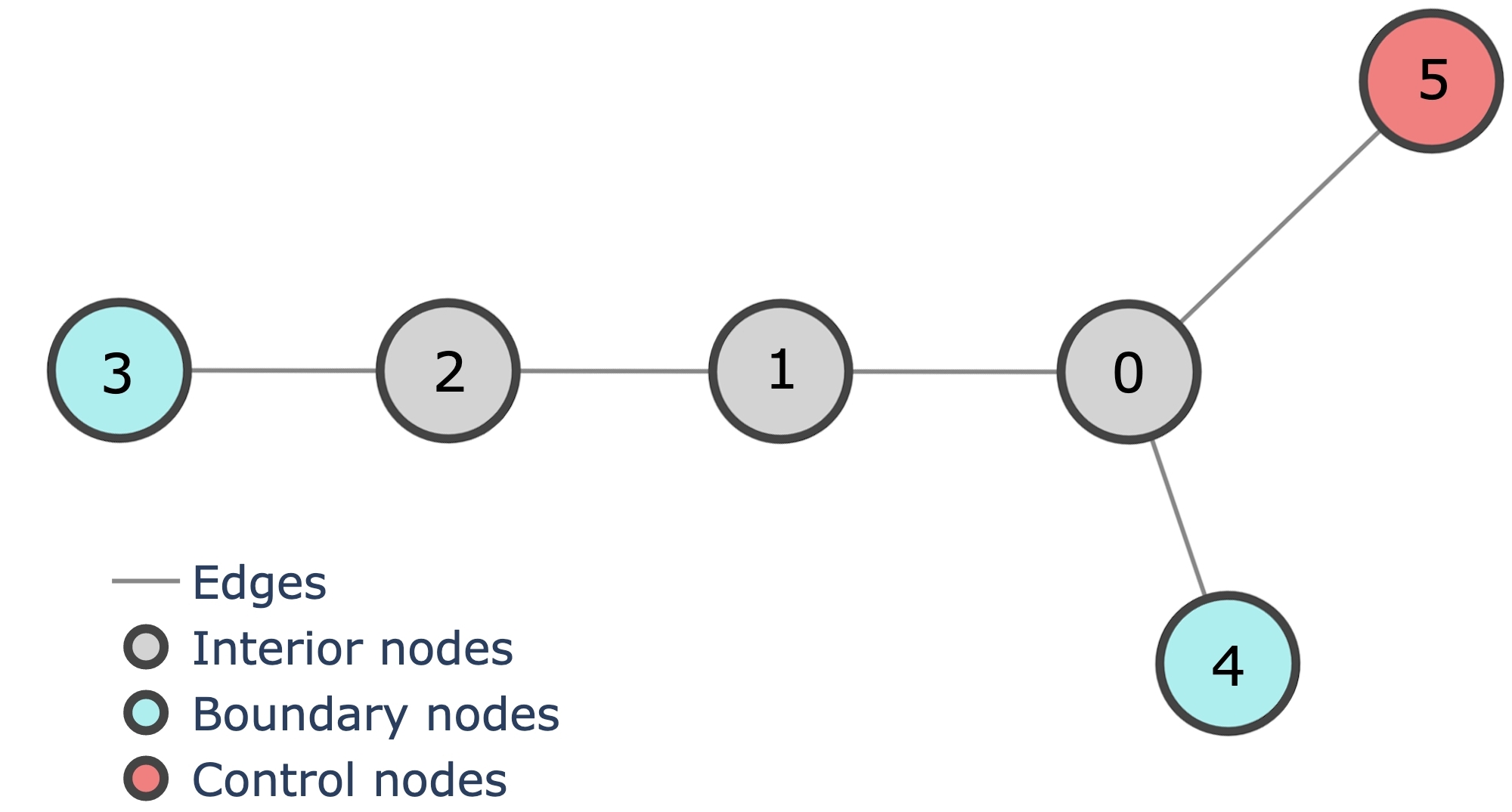}
    \caption{Graph $G_2$ used in OC experiment}
    \label{ctrl_graph}
\end{figure}

\begin{figure}\label{control_comp}
\begin{center}
 \includegraphics[scale=0.032]{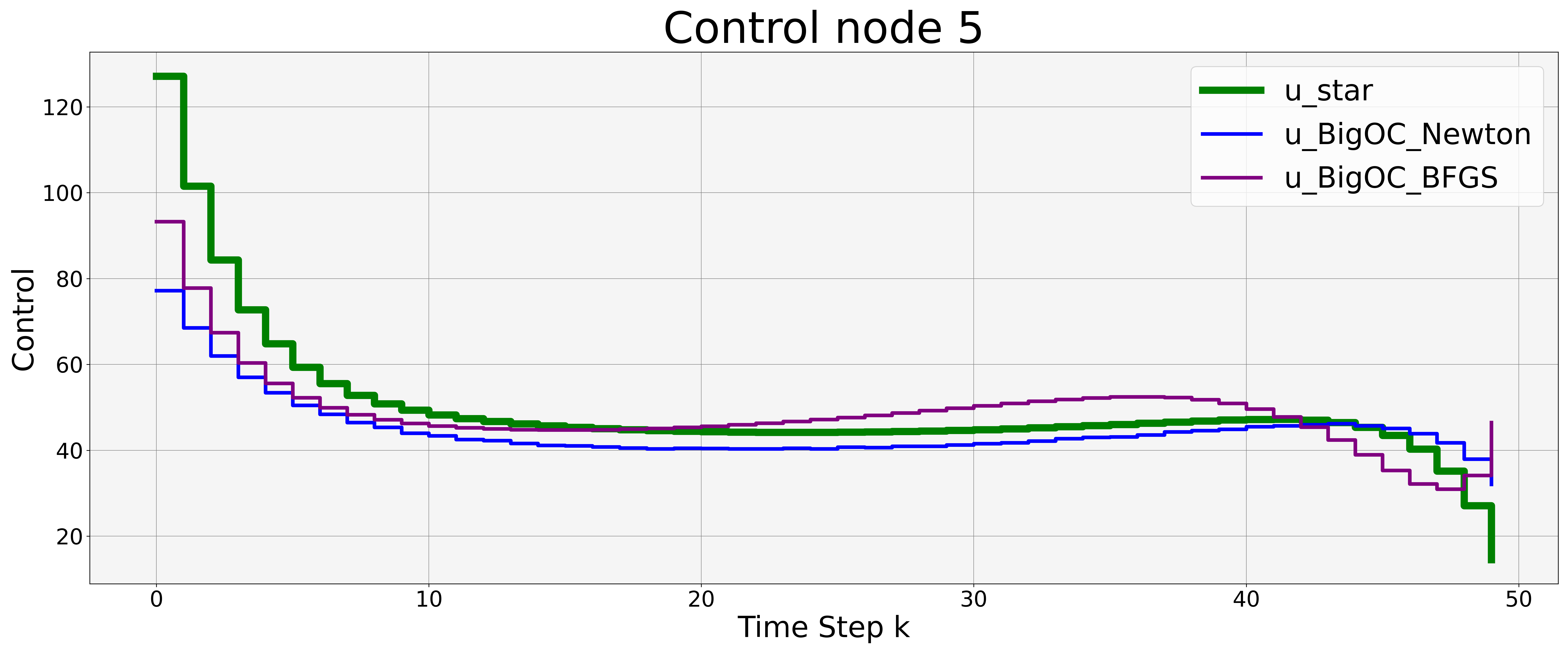}
 \caption{Comparison of two boundary injected optimal controls against true diffusion and with $N=50$}
\end{center} 
\end{figure}

In this section, we apply Algorithm \ref{main_algo} to perform optimal control on graph $G_2$ (Figure \ref{ctrl_graph}) where the $\x_{\interior G}(0)=(0, \dots, 0)$ and the boundaries are kept at zero as well, and the system dynamics is determined by the diffusion equation \ref{linear_heat} with $\kappa = 0.1$ as the true source of dynamics. We apply Casadi \cite{Andersson2019} to the source system and take the resulting optimal policy $\mathbf{u}^*$ as the ground truth. As is typical of our approach, we are unaware of the underlying differential equation. In the absence of the true differential equation, we understand the system through a \bignode model that is sufficiently trained to gain generalization power. 
We run the discrete optimal control process above for a maximum of $1000$ iterations or until the loss error drops below $10^{-6}$ with $N=50$, $dt=0.1$, $\kappa=0.5$, $\alpha=50, \beta=1, \gamma = 10^4$, and the desired state is $\hat{\x}=(16, 12, 6)$. We run this experiment using both optimization update rules, namely Newton's and BFGS, and using sensitivities. Figure \ref{control_comp} shows a comparison of our optimal controls against the ground truth and the data is organized in Table \ref{tab:ctrl_comparison}. Next, we have the following table which is a summary of the resulting errors through the interior trajectories.

\begin{table}[htbp]
  \centering
  \label{tab:diffusion_bignode_comparison}
  \begin{tabular}{l|cc|cc}
    \toprule
    dynamics & \multicolumn{2}{c|}{diffusion} & \multicolumn{2}{c}{\bignode} \\
    BigOC & Newton & BFGS & Newton & BFGS \\
    \midrule
    MAE & 1.22 & 0.68 & 0.51 & 0.63 \\
    MAPE & 16.45 & 9.51 & 26.60 & 28.20 \\
    \bottomrule
  \end{tabular}
  \caption{Comparison of mean absolute and mean average percentage errors of these trajectories relative to true diffusion}
\end{table}
Next, we want to visually compare all state trajectories produced by all combinations of dynamics and control as shown below. 
\begin{center}
 \includegraphics[scale=0.065]{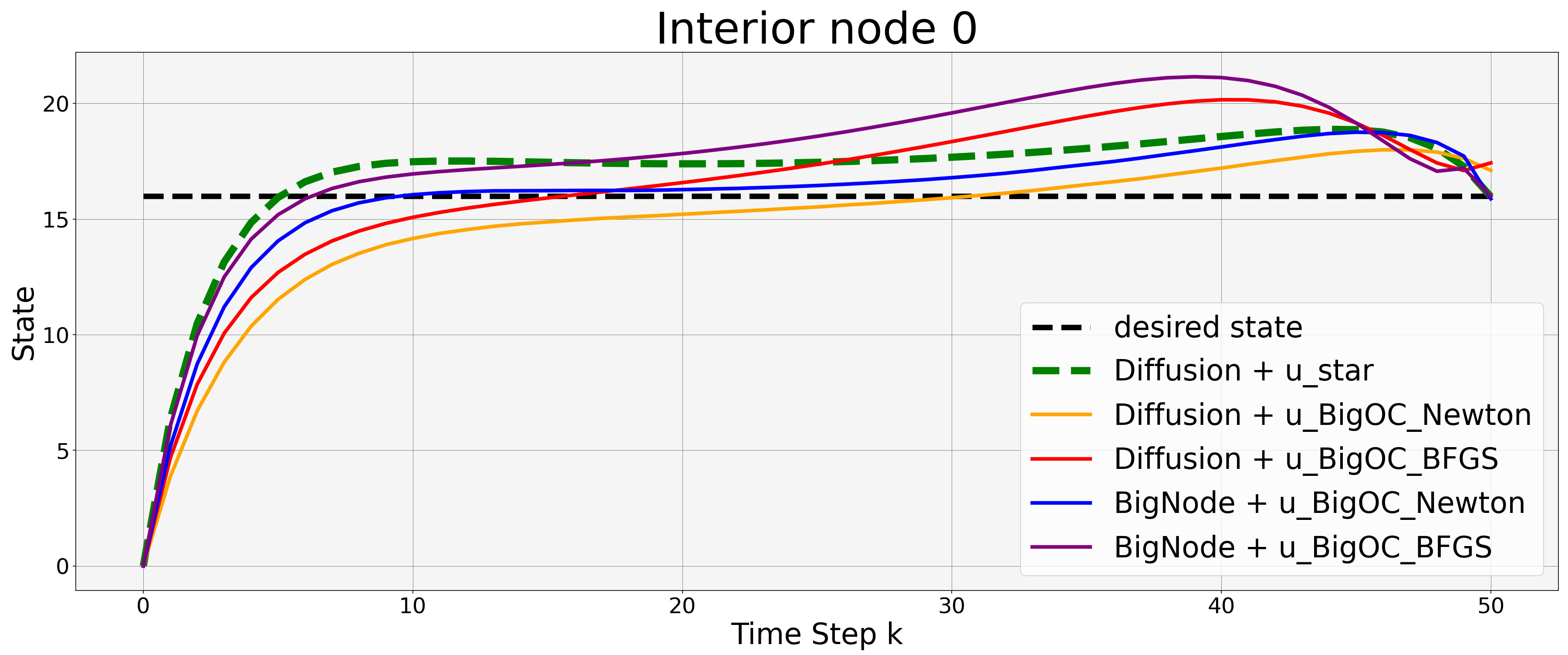}
\end{center}
\begin{center}
 \includegraphics[scale=0.065]{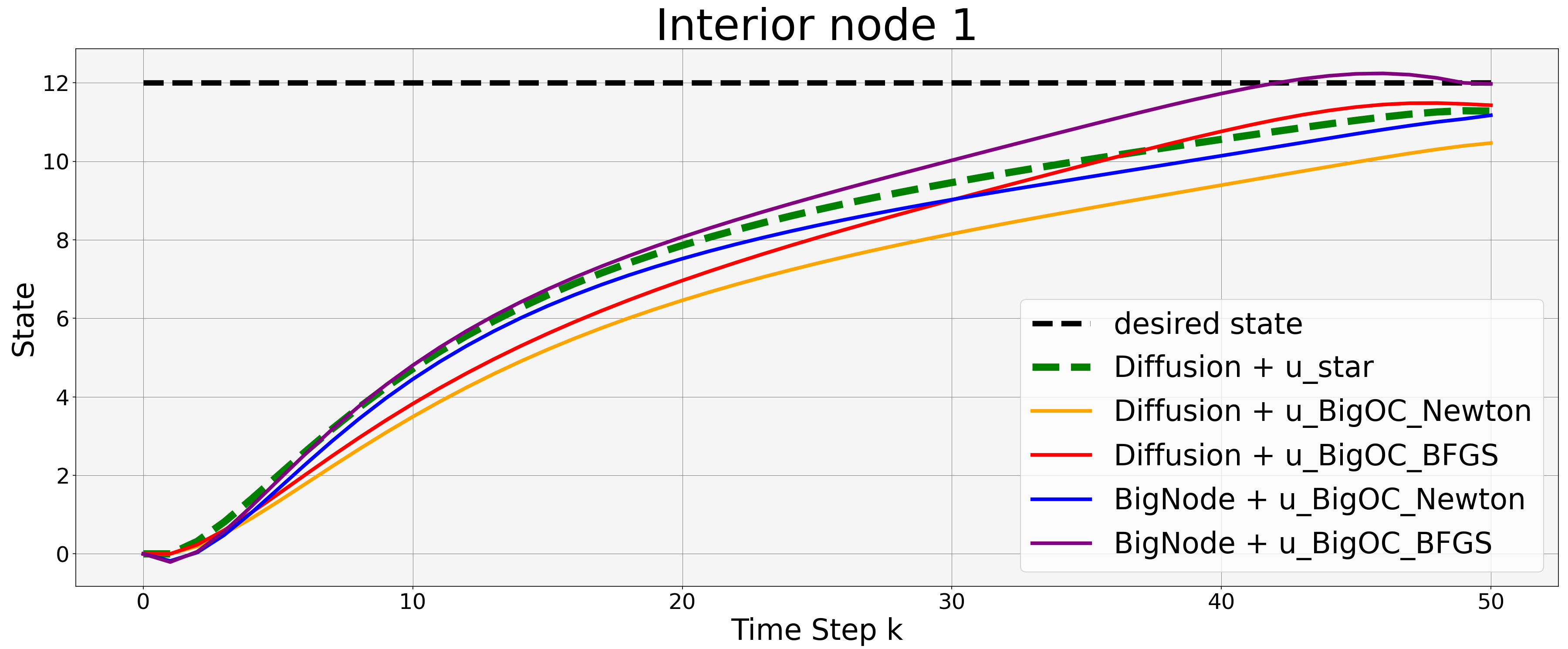}
\end{center}
\begin{center}
 \includegraphics[scale=0.065]{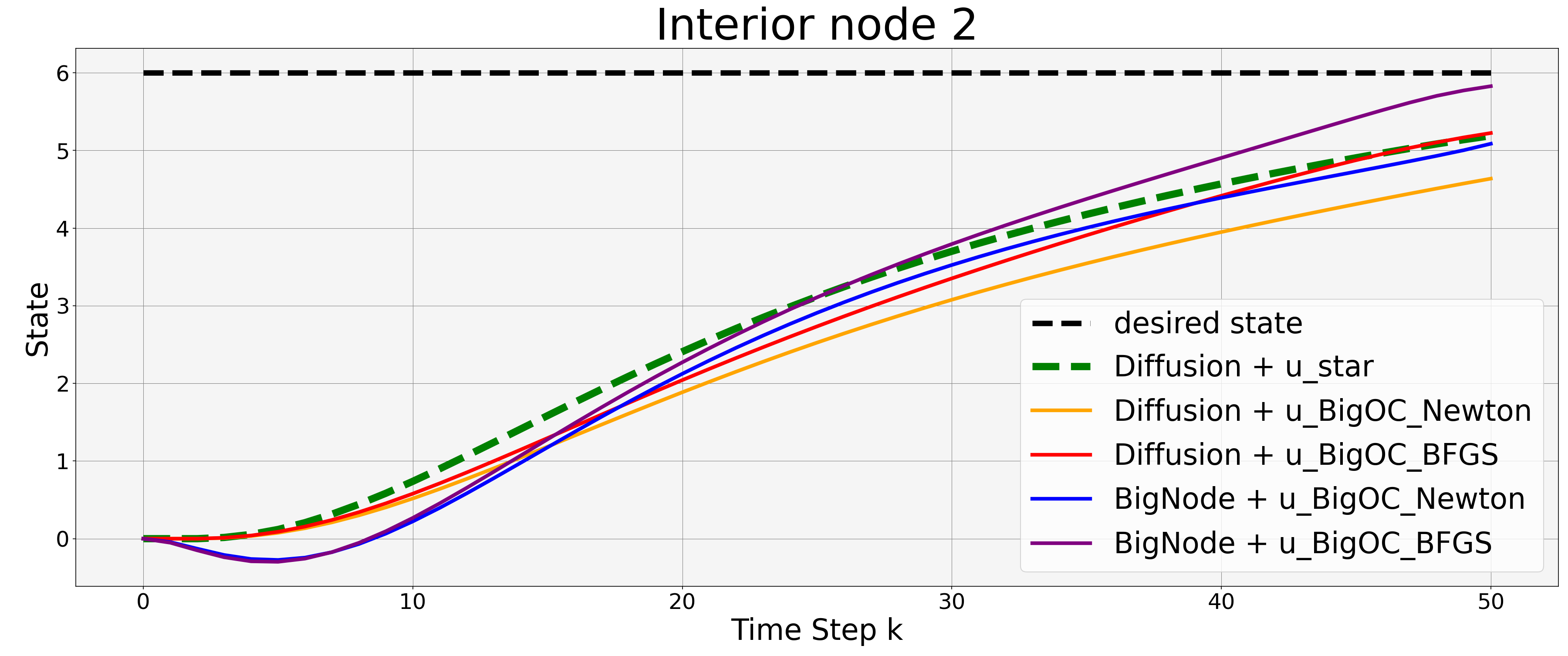}
\end{center}

\begin{table}[htbp]
  \centering
  \label{tab:ctrl_comparison}
  \begin{tabular}{lcc}
    \toprule
   Error & BigOC/Newton & BigOC/BFGS \\
    \midrule
    MAE & 6.39 & 5.92 \\
    MAPE & 12.82 & 13.95 \\
    \bottomrule
  \end{tabular}
\caption{MAE and MAPE of L1 distance between $\mathbf{u}^*$ and our controls}
\end{table}
The last table of this section is a comparison between the various control costs as in Formula (\ref{tab:ctrl_comparison}). 
\begin{table}[htbp]
  \centering
  \label{tab:diffusion_bignode_comparison}
  \begin{tabular}{l|cc|cc}
    \toprule
    dynamics & \multicolumn{2}{c|}{diffusion} & \multicolumn{2}{c}{\bignode} \\
    BigOC & Newton & BFGS & Newton & BFGS \\
    \midrule
    MAE & 12641.05 & 11028.06 & 1329.61 & 6506.30 \\
    MAPE & 3.60 & 7.57 & 3.30 & 9.48 \\
    \bottomrule
  \end{tabular}
  \caption{Comparison between optimization cost of the four combination above relative to ground truth}
\end{table}

\section{Discussions and future steps}

The focus of the present work was to develop the basics of the boundary injecction method using graph neural networks in conjunction with neural ODE solvers. This approach can be improved and extended in many directions. 

First, in regards to the system identification part, an ever present challenge is the problem of stiffness, and tackling this issue is an ongoing research program. This seems to require an efficient neural ODE regularization technique such as \cite{finlay2020train} which draws from the optimal tranport techniques. Another possible improvment is to implement a technique such as \cite{rubanova2019latent} to better handle irregularly sampled data. While in the interest of succictness, we have chosen to work with experiments of small size, in order to tackle real world problems, we need to optimize our algorithms in order to improve scalability and robustness. One possible solution that we have experimented with, is the idea of transfer learning were we learn the dynamics in a small scale and then transfer to a larger graph. 

Second, in regards to the optimal control part, the most basic improvement is to allow for specifying inequality constraints on the admissible values of the control ${\mathbf{u}}$. This is done through modifying the Lagrangian. A general challenge in tackling the optimization component is dealing with the curse of dimensionality which requires its own novel techniques. For instance, the stochastic technique introduced in \cite{hu2023tackling} could have the potential to transfer to this context to regularize optimal control. 

\section{Recap}

In this study, we established a versatile framework for system identification on graphs incorporating boundaries. This framework accommodates both external and internal dynamics, allowing for a broad range of applications. Utilizing graph neural networks in tandem with neural ordinary differential equation solvers, we conducted a series of experiments to learn these dynamics. A pivotal aspect of our architecture involves the boundary injection technique as implemented within the graph neural network.

Furthermore, we conducted experiments involving a regularization technique based on graphical distance, demonstrating its potential to enhance training quality, particularly for nodes located farther from the boundary. Leveraging this framework, when dealing with control nodes governed by a prescribed cost, we proposed an offline algorithm based on discrete optimal control. This algorithm aims to guide the dynamics toward a desired state.

Finally, it is noteworthy to highlight that a byproduct of working with graphs as the substrate is that we can emulate surfaces with nontrival topology such as discretized spheres, 2-tori, etc. 

\bibliographystyle{alpha}

\appendix
\onecolumn
\section{Appendix}

\subsection{Variations with RK4}\label{rk4}

The fourth order Runge-Kutta method is a more straightforward yet effective method to approximate $g$ when using a locally constant control $\bu_k$ in the k\textsuperscript{th}
 interval $(t_k, t_{k+1})$. For any $k=0,\dots, N-1$, let $h=t_{k+1}-t_k$ and use the following update rule based on the four slopes. 
\begin{equation}
  \begin{aligned}
 {\mathbf{k}}_1 &= hD_{\Theta}(\x_k, \bu_k, t_k),\\
 {\mathbf{k}}_2 &= hD_{\Theta}(\x_k + \frac{1}{2}k_1, \bu_k, t_k + \frac{h}{2}),\\
 {\mathbf{k}}_3 &= hD_{\Theta}(\x_k + \frac{1}{2}k_2, \bu_k, t_k + \frac{h}{2}),\\
 {\mathbf{k}}_4 &= hD_{\Theta}(\x_k+k_3, \bu_k, t_{k+1}), \\
 \x_{k+1} & = \x_k + \frac{1}{6}({\mathbf{k}}_1+2 {\mathbf{k}}_2+2 {\mathbf{k}}_3+ {\mathbf{k}}_4).
\end{aligned}
\end{equation}

\subsection{An alternative update rule: BFGS}\label{bfgs}

As is standard in numerical analysis \cite{nocedal1999numerical}, we implement the Broyden–Fletcher–Goldfarb–Shanno (BFGS) algorithm to speed up the iterative solution to the residuals. This is a process where $\mathbf{U}_l$ is updated using the formulas below to bring the value of $R(\mathbf{U})$ close to zero, starting with $\mathbf{U}_0=I$ as the identity matrix of size $N|V_{\ctrl}|$. Here, ${\mathbf{p}}_l$ is the direction of descent, and $\alpha_l$ is the step size. The update formula is:
\begin{equation}
\begin{aligned}
  {\mathbf{p}}_l & = - {\mathbf{H}}_l J_R(\mathbf{U}_l)\\
   \mathbf{U}_{l+1} & = \mathbf{U}_l + \alpha_l {\mathbf{p}}_l \\
 {\mathbf{y}}_l  & = J_R(\mathbf{U}_{l+1}) - J_R(\mathbf{U}_l), \quad
 {\mathbf{s}}_l  = \mathbf{U}_{l+1} - \mathbf{U}_{l}\\
 {\mathbf{H}}_{l+1} & = {\mathbf{H}}_l + \frac{({\mathbf{s}}_l{\mathbf{y}} + {\mathbf{y}}_l^T {\mathbf{H}}_l {\mathbf{y}}_l)({\mathbf{s}}_l{\mathbf{s}}_l^T)}{({\mathbf{s}}_l^T {\mathbf{y}}_l)^2} - \frac{ {\mathbf{H}}_l {\mathbf{y}}_l {\mathbf{s}}_l^T + {\mathbf{s}}_l {\mathbf{y}}_l^T{\mathbf{H}}_l }{{\mathbf{s}}_l^T {\mathbf{y}}_l} 
\end{aligned}
\end{equation}

\subsection{Additional plots}\label{add_plots}

Here are the complete set of trajectories of the linear experiment from Section \ref{exp1}. 
\begin{center}
\includegraphics[scale=0.45]{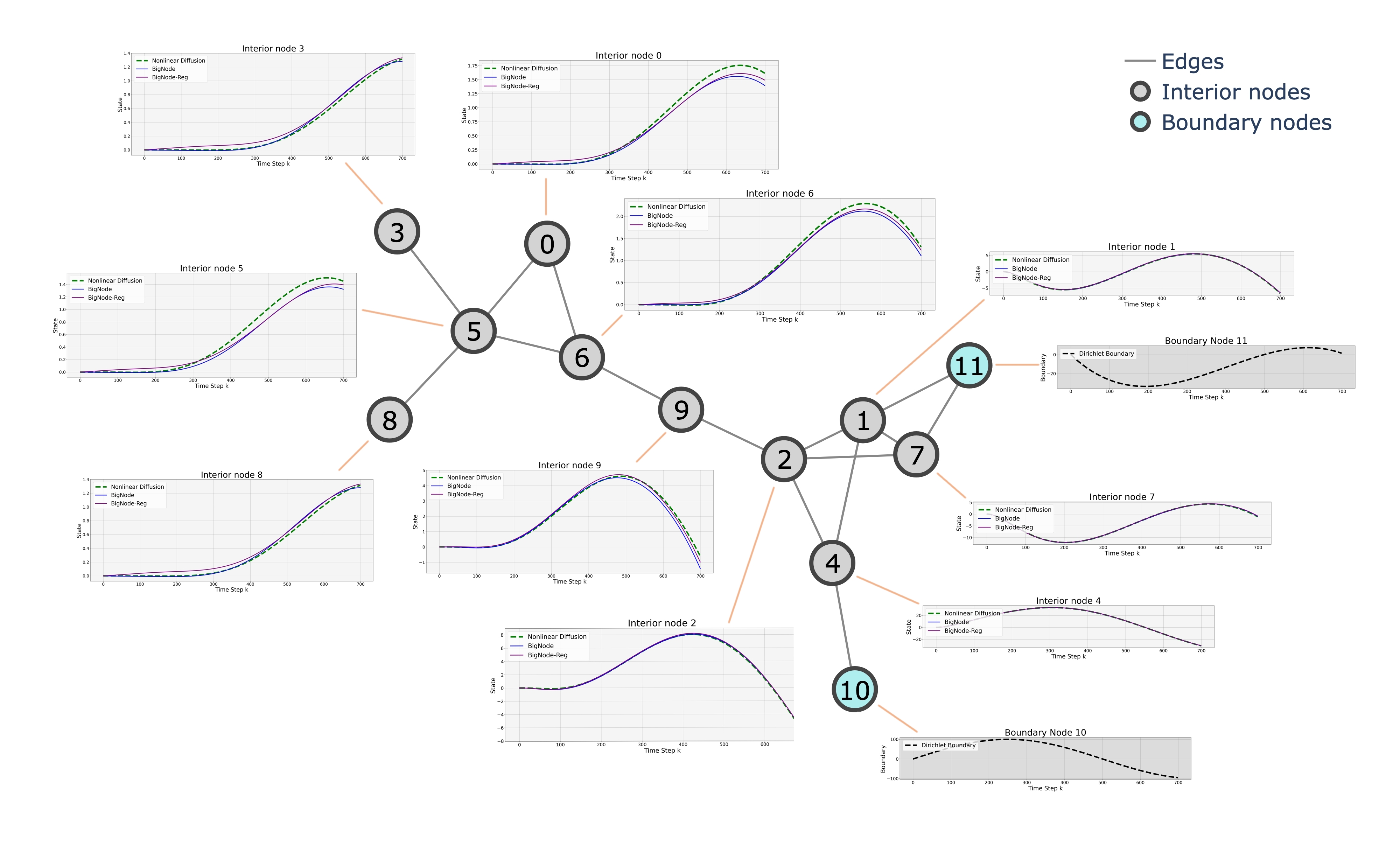} 
\end{center}

And, here are the complete set of trajectories of the nonlinear experiment from Section \ref{exp1}. 
\begin{center}
\includegraphics[scale=0.45]{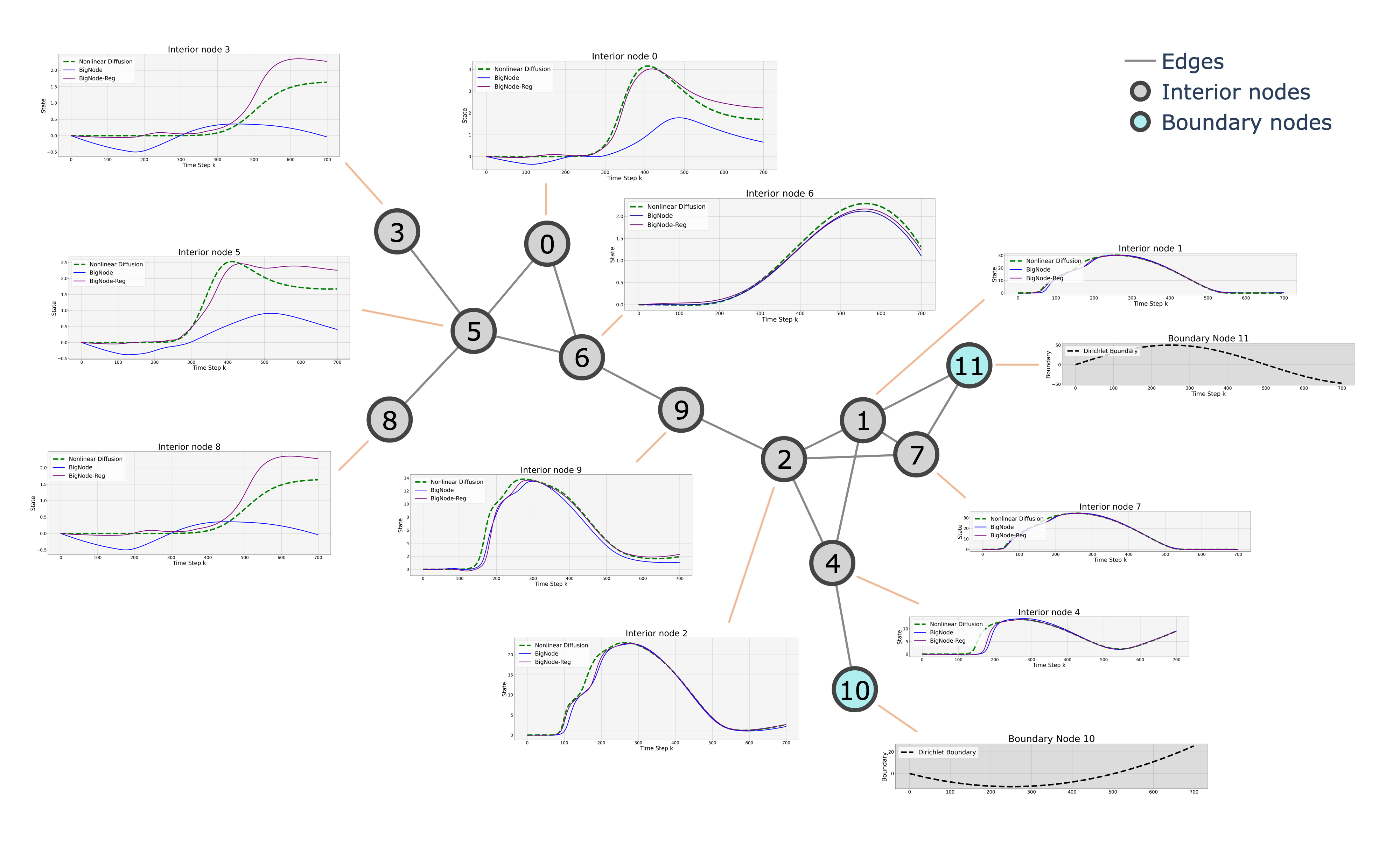} 
\end{center}

\end{document}